\pdfoutput=1

\documentclass[11pt]{article}

\usepackage{acl}

\usepackage{times}
\usepackage{latexsym}
\usepackage{amsmath,graphicx,booktabs,multicol,multirow,makecell,CJKutf8,verbatim,soul}
\usepackage[shortlabels]{enumitem}

\usepackage[T1]{fontenc}

\usepackage[utf8]{inputenc}

\usepackage{microtype}

\title{Bayesian Example Selection Improves In-Context Learning \\ for Speech, Text, and Visual Modalities}

\author{Siyin Wang$^1$, Chao-Han Huck Yang$^2$, Ji Wu$^1$, Chao Zhang$^1$ \\
  $^1$Tsinghua University, $^2$NVIDIA Research \\
  \texttt{wangsiyi23@mails.tsinghua.edu.cn,cz277@tsinghua.edu.cn} \\}

\begin{document}
\maketitle
\begin{abstract}
Large language models (LLMs) can adapt to new tasks through in-context learning (ICL) based on a few examples presented in dialogue history without any model parameter update. Despite such convenience, the performance of ICL heavily depends on the quality of the in-context examples presented, which makes the in-context example selection approach a critical choice. This paper proposes a novel \textbf{B}a\textbf{y}esian in-\textbf{C}ontext example \textbf{S}election method (ByCS) for ICL. 
Extending the inference probability conditioned on in-context examples based on Bayes' theorem, ByCS focuses on the inverse inference conditioned on test input. Following the assumption that accurate inverse inference probability (likelihood) will result in accurate inference probability (posterior), in-context examples are selected based on their inverse inference results. Diverse and extensive cross-tasking and cross-modality experiments are performed with speech, text, and image examples. Experimental results show the efficacy and robustness of our ByCS method on various models, tasks and modalities.

\end{abstract}

\section{Introduction}
Large language models (LLMs) \citep{llama2,gpt4} have achieved great success on many text-based natural language processing (NLP) tasks. By connecting with extra visual and audio encoders \citep{eva-clip,whisper}, the resulting multimodal LLMs can also achieve remarkable performance on image-text and audio-text tasks \cite{blip2,gpt4v,salmonn}. With the ability of in-context learning (ICL) \citep{gpt3}, LLMs can adapt to new tasks easily and efficiently in a training-free manner, to generate output following the prompting paradigm based on a few input-label pairs pre-pended to the test input. The existence of ICL ability has also been verified on image-text and audio-text tasks \citep{frozen,sicl,hsu2023gslm,pan2023cosmic}.

\begin{figure}[h]
\centering
\includegraphics[width=3.1in]{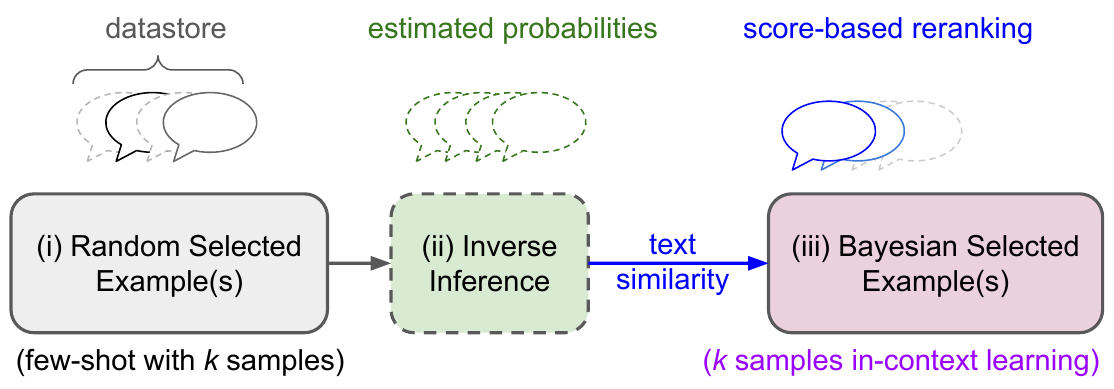}
\vspace{-0.4cm}
\caption{A brief illustration of the proposed Bayesian in-context example selection includes: (i) first randomly selecting $k$ examples; (ii) examining the examples in the datastore through ``inverse inference,'' where the test input-label pair serves as the in-context example; and (iii) selecting samples with correct label predictions as good examples (colored in \textcolor{blue}{blue}), considered to have high mutual information interaction with the test input.}
\label{fig:1:all}
\end{figure}

Although ICL requires no gradient descent and thus does not suffer from the instability caused by stochastic optimisation compared to other test-time adaptation approaches, care still needs to be taken when selecting the in-context examples since they often lead to distinct ICL performance variations \citep{zhao2021calibrate,min2022rethinking,lu2022fantastically}. 
Prior work on in-context example selection trains an example retrieval module \citep{rubin2022,zhang2022active,lu2022dynamic,wang2023learning}, selects close examples in embedding space \citep{liu2022knn,an2023skill,qin2023ids}, or leverages the feedback of LLMs to score the examples \citep{su2022selective,nguyen2023influence,iter2023ced,mavromatis2023ada}. While boosting ICL performance, most methods treat in-context examples and test input separately, overlooking their mutual interactions.

\begin{figure}[h]
\centering
\includegraphics[width=0.48\textwidth]{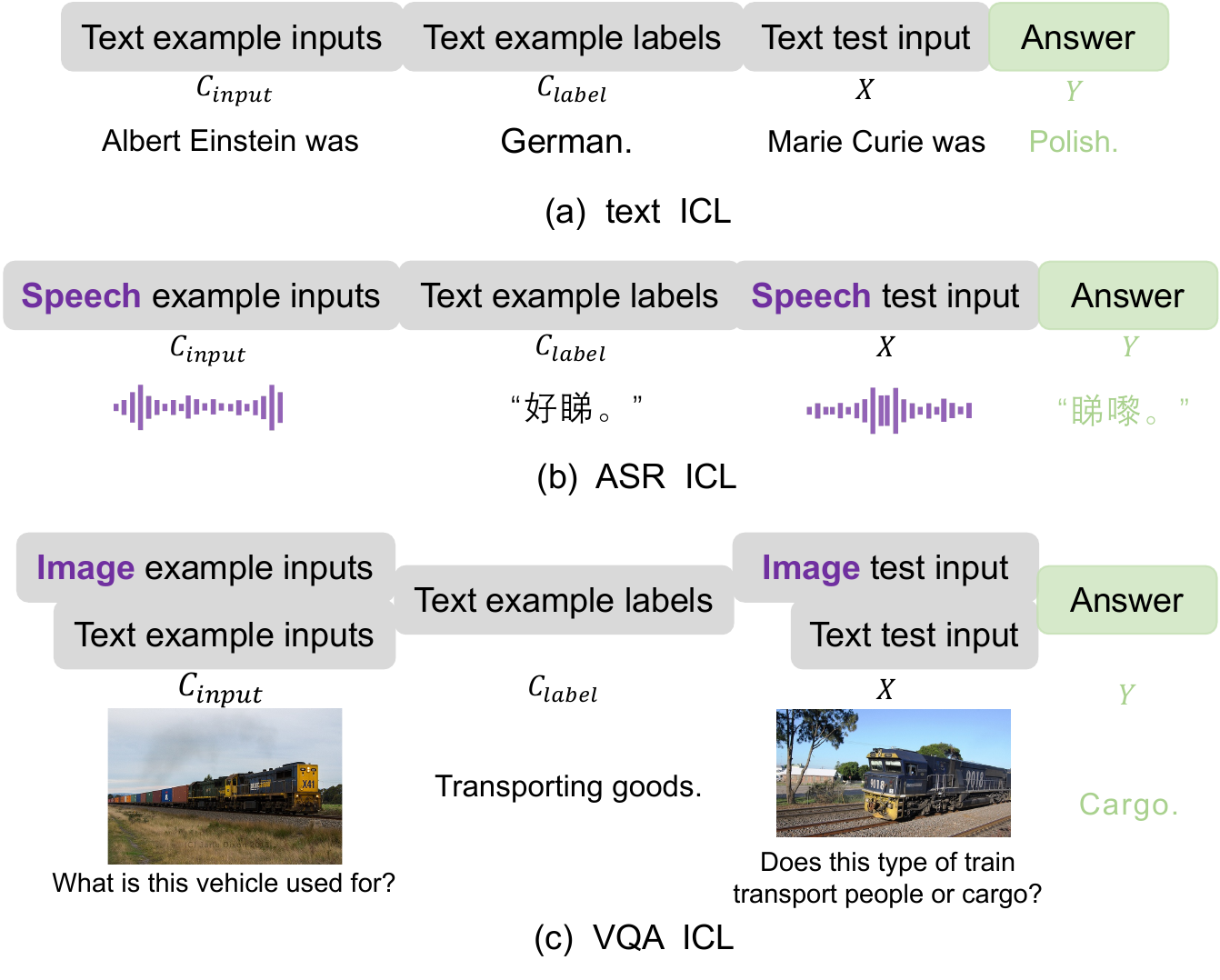}
\caption{Multimodal ICL. Although ICL on different modalities shares the same formula expression, the actual inputs and inference model architectures differ. For ASR ICL on Whisper, the speech is fed into the encoder while the text example is labelled into the decoder, which is aware of speech input through cross-attention with the encoder. For VQA ICL, images are first encoded to the same embedding space of LM's input, then interleaved images and texts are fed into decoder LM.}
\label{figure:2}
\end{figure}

This paper proposes ByCS (\textbf{B}a\textbf{y}esian in-\textbf{C}ontext example \textbf{S}election), a novel in-context example selection approach focusing on mutual information interactions based on the Bayesian formula. Refer to the inference of test input conditioned on in-context examples as ICL \textit{inference}, and the inference of in-context example's input based on the test input-label pair as the \textit{inverse inference}. By introducing inverse inference via \textit{Bayes' theorem}, ByCS leverages the inverse inference result to evaluate the quality of each in-context example. Assuming the contextual information interaction is mutual, an accurate inverse inference is likely to result in an accurate inference. Examples with accurate inverse inference results are selected as optimal examples.
Extensive experiments across audio, image, and text modalities are conducted to verify the effectiveness and robustness of ByCS, such as ASR, visual question answering (VQA), as well as NLP tasks (including topic classification, sentiment analysis, and text-to-SQL \textit{etc}). Our main contributions are summarised as follows:
\begin{itemize}
    \item ByCS, a novel in-context example selection method inspired by Bayes' theorem, is proposed. To improve the efficiency, the use of a smaller model for fast inverse inference implementation and a ranking-based pre-selection to reduce the number of in-context examples are also proposed in this paper.  
    \item The method is verified using both ``decoder-only ICL" on NLP tasks and ``encoder-decoder'' ICL on ASR and VQA. To the best of our knowledge, this is the first work of an in-context example selection method verified across text, audio, and visual modalities as shown in Figure~\ref{figure:2}. 
\end{itemize}

\section{Related Work}


\textbf{Multimodal ICL.} Inspired by the decoder-only ICL in text-based NLP, efforts have been made to extend such a few-shot learning ability to other modalities, in particular image and audio. Frozen \citep{frozen} is the first attempt to exploit ICL ability in the vision-language model (VLM). By using a vision encoder to map the input image to textual tokens in the input embedding space of a frozen text language model, Frozen can handle interleaved image and text input and achieve image-text ICL. Other work manages to improve VLM's ICL ability by using adapter blocks \citep{magma}, adding blockwise modality fusion structures \citep{flamingo} and scaling up the model size \citep{emu2}.


In audio modality, \citet{audioLM} proposed AudioLM, a language model based on quantised audio tokens for audio generation tasks, which exhibits ICL ability for audio continuation. Similarly, \citet{valle} proposed  VALL-E, a controllable text-to-speech synthesis system with ICL ability based on audio and text prompts. 
\citet{sicl} presented the first ICL work for ASR based on paired speech-text examples, which adapted the Whisper \citep{whisper} model to receive considerable word error rate (WER) reductions on unseen Chinese dialects. Further explorations enabled the recent speech-language models to perform ICL on more speech input tasks through warmup training \citep{hsu2023gslm} or speech instruction-tuning \citep{pan2023cosmic}.

\begin{figure*}[t]
\centering
\includegraphics[width=0.85\textwidth]{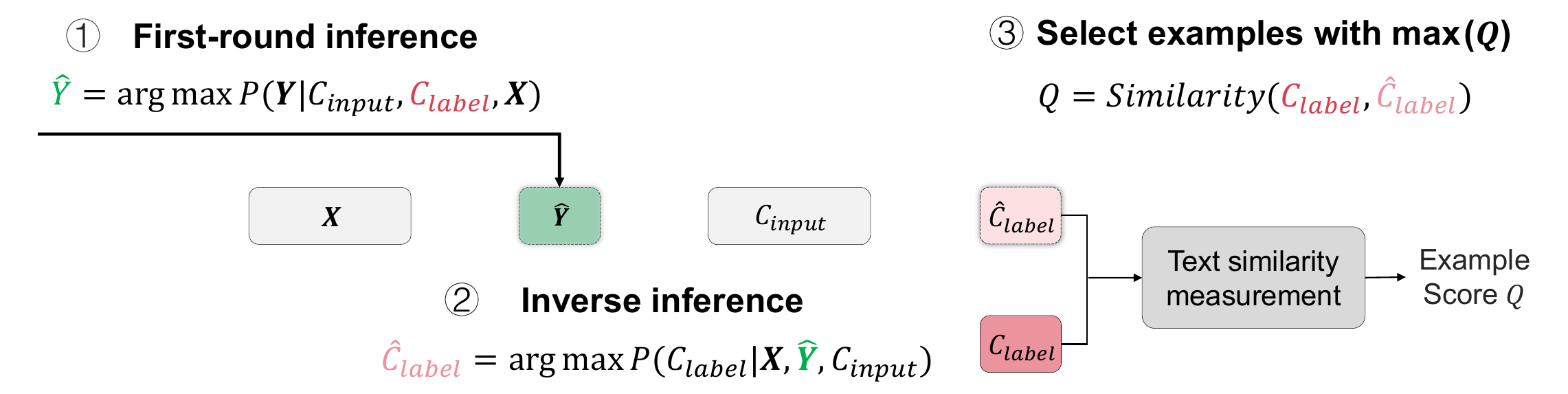}
\vspace{-0.1cm}
\caption{The detailed pipeline of our ByCS method includes: First, conduct the first-round inference to estimate the label of the test input. Then, perform inverse inference on each example in the datastore, where the test input and the estimated label serve as in-context examples. A detailed illustration of inverse inference can be found in Figure \ref{fig:5} in the Appendix. Finally, rank in-context examples by the text similarity between the inverse inference result and the true context label. Examples with high similarity scores are selected due to their high mutual information interaction.}
\label{figure:3}
\end{figure*}

\noindent
\textbf{In-Context Example Selection Methods.} 
\citet{rubin2022} proposed a scoring LM to retrieve in-context examples using contrastive learning, which can also be trained with reinforced learning algorithms, such as Q-learning \citep{zhang2022active} and policy gradient \citep{lu2022dynamic}. Alternatively, examples that are semantically similar to the test input can be selected. \citet{liu2022knn} proposed to select the $k$ nearest neighbours ($k$NN) in the embedding space of the examples. When combining with chain-of-thought \citep{wei2022cot}, \citet{qin2023ids} proposed to select examples in the embedding space of the reasoning path.
LLM feedback is often used in in-context example selection. 
\citet{iter2023ced} selected in-context examples with cross-entropy differences of the fine-tuned model based on the assumption that ICL may act as implicit gradient descent \citep{dai2022gradient}. \citet{nguyen2023influence} identified highly impactful examples according to the proposed influence score. Although ByCS also uses LLM feedback when evaluating the quality of in-context examples through inverse inference, it leverages the text-similarity of the inverse inference results and the corresponding ground-truth labels, in no need of complete output probability distributions which are often not available for commercial LLMs.



\citet{wang2023latent} selected optimal in-context examples in the Bayesian framework by viewing LLMs as latent variable models and ICL as latent concept learning. In comparison, ByCS directly extends the ICL inference probability using Bayes' theorem. \citet{xu2024misconfidence} selected examples with high discrepancy between the labels and LLM's outputs when performing question answering. ByCS also selected examples from candidates in a datastore based on LLM's outputs but computes the mutual information interactions between the in-context examples and test input.

\section{Methodology}
As shown in Figure~\ref{figure:3}, given a test input $\mathbf{X}$ and paired in-context examples $(\mathcal{C}_\text{input}, \mathcal{C}_\text{label})$, LLMs predict the most possible answer $\hat{\mathbf{Y}}$ by maximising the inference probability $P(\mathbf{Y}|\mathcal{C}_\text{input},\mathcal{C}_\text{label},\mathbf{X})$: 
\begin{equation}
\label{eq:1}
\hat{\mathbf{Y}} =\arg\max P(\mathbf{Y}|\mathcal{C}_\text{input},\mathcal{C}_\text{label},\mathbf{X}),
\end{equation}
where $\mathcal{C}_\text{input}$ and $\mathcal{C}_\text{label}$ are the inputs and labels of different data types in different tasks. Regarding text-based NLP tasks, $\mathcal{C}_\text{input}$ and $\mathcal{C}_\text{label}$ are referred to as text questions and corresponding answers. Regarding ASR, $\mathcal{C}_\text{input}$ and $\mathcal{C}_\text{label}$ are speech audio and corresponding text transcriptions. Regarding VQA, $\mathcal{C}_\text{input}$ are images and text questions based on the images and $\mathcal{C}_\text{label}$ are the text answers.

The inference probability can be extended using Bayes' theorem:
\begin{equation}
\begin{aligned}
P&(\mathbf{Y}|\mathcal{C}_\text{input},\mathcal{C}_\text{label},\mathbf{X}) \\ &=
\frac{P(\mathcal{C}_\text{label}|\mathbf{X},\mathbf{Y},\mathcal{C}_\text{input})P(\mathbf{Y}|\mathbf{X},\mathcal{C}_\text{input})}{P(\mathcal{C}_\text{label}|\mathbf{X},\mathcal{C}_\text{input})}.
\end{aligned}
\end{equation}

The likelihood $P(\mathcal{C}_\text{label}|\mathbf{X},\mathbf{Y},\mathcal{C}_\text{input})$ is termed as  \textit{inverse inference probability}, since it can be interpreted as the probability of the context label $\mathcal{C}_\text{label}$ when the test input-label pair $(\mathbf{X},\mathbf{Y})$ is inversely treated as the in-context example. ByCS is focused on the inverse inference probability and assumes the influence of the prior $P(\mathbf{Y}|\mathbf{X},\mathcal{C}_\text{input})$ is subordinate for simplification. 

In practice, since the ground-truth label ${\mathbf{Y}}_\text{ref}$ of the test input $\mathbf{X}$ is not available, the correct likelihood $P(\mathcal{C}_\text{label}|\mathbf{X},\mathbf{Y}_\text{ref},\mathcal{C}_\text{input})$ is approximated by $P(\mathcal{C}_\text{label}|\mathbf{X},\hat{{\mathbf{Y}}},\mathcal{C}_\text{input})$, where $\hat{{\mathbf{Y}}}$ is produced by the first-round inference. Specifically, 
\begin{itemize}
\item First, the first-round inference is performed to produce a hypothesized label $\hat{{\mathbf{Y}}}$ based on the test input $\mathbf{X}$, which can be achieved using decoding rule without any in-context examples by $\hat{{\mathbf{Y}}}=\arg\max P(\mathbf{Y}|\mathbf{X})$. Better performance can be achieved when using the hypothesized label obtained by in-context examples by $\hat{{\mathbf{Y}}}=\arg\max P(\mathbf{Y}|\tilde{\mathcal{C}}_\text{input},\tilde{\mathcal{C}}_\text{label},\mathbf{X})$ based on Eqn.~\eqref{eq:1}, where $(\tilde{\mathcal{C}}_\text{input},\tilde{\mathcal{C}}_\text{label})$ is a pair of first-round in-context example selected either randomly or using other example selection methods. 

\item Next, for the datastore with all candidate in-context examples, generate the inverse inference result in $\hat{\mathcal{C}}_\text{label}$ for every candidate example based on the approximated inverse inference probability $P(\mathcal{C}_\text{label}|\mathbf{X},\hat{{\mathbf{Y}}},\mathcal{C}_\text{input})$ by $\hat{\mathcal{C}}_\text{label} =\arg\max P(\mathcal{C}_\text{label}|\mathbf{X},\hat{{\mathbf{Y}}},\mathcal{C}_\text{input})$. 
\item Last, compute $\mathcal{Q}=\text{Similarity}({\mathcal{C}}_\text{label},\hat{\mathcal{C}}_\text{label})$ as the text similarity between ${\mathcal{C}}_\text{label}$ and $\hat{\mathcal{C}}_\text{label}$, and use $\mathcal{Q}$ as the metric for the evaluation of the quality of inverse inference.  
Since more accurate inverse inference probability often results in higher text similarity, ByCS selects the in-context examples with higher $\mathcal{Q}$.
Note that $\mathcal{Q}$ is adopted since it does not require to assessment of the model's output probability distribution of the LLM, which is often unavailable for commercial LLMs.  

\end{itemize}

To reduce the computation cost of inverse inference, two methods are used when the number of examples in the datastore is large: 
\begin{itemize}
\item Conduct inverse inference using a model in the same model family as our inference model but has a smaller model size.
\item Apply ByCS to a small number (\textit{e.g.} $N$) of pre-selected candidate examples. In pre-selection, all examples in the datastore are first ranked, and only the top $N$ best examples are reserved as the pre-selected candidates. The pre-selection is performed using fast ranking-based algorithms like $k$NN.  
\end{itemize}

\section{Experimental Setup}

\subsection{Models}
Experimental results are performed on audio, text, and image modalities. For audio-text and image-text tasks, ASR and VQA are used to evaluate the ICL ability of encoder-decoder structured models. For text-only NLP tasks, topic classification, sentiment analysis, and text-to-SQL are used to evaluate the ICL performance with decoder-only models. 
Regarding the NLP tasks, experiments are conducted using GPT-3.5-Turbo and GPT-4 \citep{gpt4}. For the ASR task, the open-sourced Whisper model \citep{whisper} is used, which is a series of speech models released by OpenAI. The Whisper model family uses vanilla encoder-decoder Transformer \citep{transformer} architecture ranging from 39 million (M) parameters (tiny) to 1.55 billion (B) parameters (large). Specifically, the Whisper small (244M) and Whisper large-v2/-v3 (1.55B) models are used. For the VQA task, experiments are performed on Emu2 \citep{emu2} and GPT-4V \citep{gpt4v}. Emu2 is a 37B text-image model (VLM) which leverages pre-trained EVA-02-CLIP-E-plus \citep{eva-clip} and LLAMA-33B \citep{llama}, which has ICL ability when taking interleaved inputs of images and texts. For experiments on Emu2, the outputs are generated using a greedy decoding setting for fast evaluation. GPT-4V is a GPT4 variant that can directly perceive image inputs, showing state-of-the-art image understanding performance.

\subsection{Datasets}

Seven datasets covering NLP, ASR and VQA are used in this paper. For text-only ICL, four datasets are used in four different task categories: the TREC dataset for topic classification \citep{trec}, the SST2 dataset 
for sentiment analysis \citep{sst5}, the Spider dataset for text-to-SQL \citep{spider}, and the CHiME-4 \citep{chime4} split of the HyPoradise dataset \citep{hyporadise} for generative language model re-scoring to correct pre-generated ASR transcriptions. 
For audio-text ICL, Two datasets are used for ASR tasks, namely RASC863 \citep{863} and CORAAL \citep{coraal}. RASC863 is a commonly used Chinese dialect ASR dataset and its dialectal words split of Chongqing and Guangzhou dialects are used. CORAAL is an English corpus with speech recordings from regional African Americans. For image-text ICL, VQA experiments are conducted on OKVQA \citep{okvqa}, a dataset that requires methods to draw upon external knowledge to answer the visual questions.

\begin{table*}[h]
\setlength{\tabcolsep}{3pt}
\centering
\begin{tabular}{c|cccc|cccc|c}
\toprule
 & \multicolumn{9}{c}{Corpus \& In-context example number $k$} \\ [0.05cm]
Setting & \multicolumn{4}{c|}{RASC863 Chongqing} & \multicolumn{4}{c|}{RASC863 Guangzhou} & CORAAL <15s\\ [0.05cm]
 & $k=1$ & $k=2$ & $k=3$ & $k=4$ & $k=1$ & $k=2$ & $k=3$ & $k=4$ & $k=1$ \\
\midrule
random & 67.1 & 56.1 & 52.7 & 51.0 & 61.7 & 38.3 & 31.2 & 28.8 & 13.2\\ 
KATE+ & 67.1 & 54.7 & 51.3 & 49.7 & 61.3 & 36.1 & \textbf{26.9} & \textbf{24.8} & 12.6\\ 
ByCS & \textbf{62.4} & \textbf{53.4} & \textbf{50.6} & \textbf{48.6} & \textbf{49.5} & \textbf{31.9} & 27.1 & 26.6 & \textbf{12.4}\\ 
\textcolor{gray}{oracle ByCS} & \color{gray}62.4 & \color{gray}52.4 & \color{gray}49.5 & \color{gray}47.2 & \color{gray}49.4 & \color{gray}30.7 & \color{gray}25.8 & \color{gray}24.7 & \color{gray}12.4\\
\bottomrule
\end{tabular}

\vspace{0.15cm}
(a) Results with Whisper-large-v2
\vspace{0.15cm}

\begin{tabular}{c|cccc|cccc|c}
\toprule
 & \multicolumn{9}{c}{Corpus \& In-context example number $k$} \\ [0.05cm]
Setting & \multicolumn{4}{c|}{RASC863 Chongqing} & \multicolumn{4}{c|}{RASC863 Guangzhou} & CORAAL <15s\\ [0.05cm]
 & $k=1$ & $k=2$ & $k=3$ & $k=4$ & $k=1$ & $k=2$ & $k=3$ & $k=4$ & $k=1$ \\
\midrule
random & 68.9 & 60.3 & 57.0 & 55.7 & 67.1 & 42.8 & 38.3 & 35.2 & 12.4\\ 
KATE+ & 68.1 & 58.2 & 54.8 & 54.1 & 67.7 & 41.3 & 34.3 & 31.6 & 12.1\\ 
ByCS & \textbf{63.5} & \textbf{56.3} & \textbf{53.5} & \textbf{51.8} & \textbf{50.7} & \textbf{36.7} & \textbf{33.0} & \textbf{31.5} & \textbf{12.0}\\ 
\color{gray}oracle ByCS & \color{gray}63.4 & \color{gray}55.2 & \color{gray}53.0 & \color{gray}50.7 & \color{gray}51.3 & \color{gray}35.6 & \color{gray}31.9 & \color{gray}30.7 & \color{gray}11.9\\
\bottomrule
\end{tabular}

\vspace{0.15cm}
(b) Results with Whisper-large-v3
\vspace{-0.15cm}
\caption{\%WERs on RASC863 dialectal word dataset and CORAAL with different in-context example selection methods. For RASC863, the example datastore is the RASC863 dialectal word dataset of the corresponding dialect. For CORAAL, the size of the example datastore for ByCS is narrowed down to 10 using $k$NN algorithm. For the ``oracle ByCS'' setting, the ground-truth label $\mathbf{Y}_\text{ref}$ is used in the inverse reference.}
\label{t1}
\end{table*}

\subsection{Baselines}

On all three modalities, \textbf{random selection} and improved \textbf{KATE} \citep{liu2022knn} are used as baseline approaches. For random selection, in-context examples are uniformly selected from the example datastore three times and the average results are reported. For KATE \citep{liu2022knn}, $k$ neighbours that are nearest to the test input in the embedding space in terms of Euclidean distance are selected. For ASR ICL, the encoder of Whisper large-v2 acts as the embedding retrieval module on the Chinese dataset, while on the English dataset, we use the encoder of Whisper large-v3. In text-ICL, OpenAI \texttt{text-embedding-ada-002} is used as the embedding retrieval model. For VQA ICL, KATE is only based on the embedding space of the query image and EVA02-CLIP-bigE-14-plus \citep{eva-clip} serves as the embedding retrieval module. We use the term “KATE+” to refer to the baseline in our paper, putting stress on the fact that it is actually an improved KATE version enhanced using stronger embedding retrieval models, which results in better performance. For text ICL, \textbf{bm25} \citep{okapi} and \textbf{LLM-R} \citep{wang2023learning} are also compared as baselines. bm25 is a ranking metric originally designed for search engines to estimate the relevance of documents to a given query based on word-overlapping similarity. LLM-R provides a recent and preferment dense retriever distilled using a reward model trained based on LLM feedback.

\section{Results}

\subsection{ASR ICL}


Results in WER are reported for ASR tasks in Table \ref{t1}, and here in Chinese WER is calculated based on Chinese characters, which is also termed as character error rate.


The ByCS method outperforms the KATE+ baseline in most cases, showing the robustness and effectiveness of our method. When the number of in-context examples $k$ is small, ByCS surpasses KATE+ baseline in a large margin, with a 10.25\% relative WER reduction on average when $k=1$. Such performance advantage of ByCS reduces when the number of in-context examples increases, which may be attributed to the fact that ByCS performs the inverse inference of each in-context example individually by applying an independence assumption that ignores the contextual interactions between different in-context examples. 
The use of $\mathbf{Y}_\text{ref}$ in ``oracle ByCS'' further boosts the performance gain, indicating the upper bound of our method with the same number of $k$.

\subsection{Ablation study on ASR ICL}

\subsubsection{Inverse decoding option}

\begin{CJK*}{UTF8}{gkai}
The influence of different decoding options of inverse inference is studied on the RASC863 dialectal word dataset. The results are shown in Table \ref{t2}.
For the setting notation, ``noprompt'' denotes decoding in the default decoding option, and ``prompt'' means to decode with a specially designed \textit{prompt} ``识别方言'' (meaning to ``recognize dialect speech''). ``LID'' denotes decoding with the correct language identity of Chinese (``zh'').
\end{CJK*}


The results show that among the three inverse decoding options, ``noprompt'' obtains the best performance, ``prompt'' becomes the second, and ``LID'' the worst. The WERs of inverse inference are reported in Table \ref{t3}. The WERs under the ``noprompt'' setting are more than 100\% due to the high insertion error rate. The repeated outputs are not removed when calculating the WERs of inverse inference and when calculating the text similarity, making a more obvious distinction between the examples with high mutual information interaction and those with low.


Although it may be a little counter-intuitive that low inverse inference accuracy results in high ByCS selection performance, it is reasonable since inverse inference in ByCS helps to separate good in-context examples from the rest, which can be better achieved by using worse decoding options during inverse inference. This is because our decoding options can often make the model make more mistakes for worse in-context examples.



\begin{table}[h]
\vspace{0.1cm}
\setlength{\tabcolsep}{3pt}
\centering
\resizebox{0.5\textwidth}{!}{
\begin{tabular}{cc|cc}
\toprule
\multicolumn{2}{c|}{Setting} & \multicolumn{2}{c}{Corpus} \\ [0.05cm]
Text  & Inverse & \multirow{3}{*}{\makecell{RASC863 \\ Chongqing}} & \multirow{3}{*}{\makecell{RASC863 \\ Guangzhou}} \\ 
similarity & decoding & & \\
measurement & option & & \\
\midrule
\multirow{3}{*}{\makecell{Jaccard \\ coefficient}} & noprompt & \textbf{62.4} & \textbf{49.5} \\ 
 & prompt & 62.9 & 50.7  \\ 
 & LID & 64.1 & 52.3 \\ 
 \midrule
\multirow{3}{*}{\makecell{BERT \\ wordvecs}} & noprompt & \textbf{62.4} & 51.5 \\
 & prompt & 63.5 & 56.8  \\ 
 & LID & 64.5 & 57.7  \\ 
\bottomrule
\end{tabular}
}
\caption{\%WERs of Whisper large-v2 on RASC863 dialectal word dataset using ByCS method with different inverse decoding options and text similarity measurements. The number of in-context examples is $k=1$.}
\label{t2}
\end{table}

\begin{table}[h]
\vspace{0.1cm}
\setlength{\tabcolsep}{3pt}
\centering
\begin{tabular}{c|cc}
\toprule
\multirow{2}{*}{\makecell{Inverse \\ decoding \\ option}} & \multicolumn{2}{c}{Corpus} \\ [0.05cm]
 & \makecell{RASC863 \\ Chongqing} & \makecell{RASC863 \\ Guangzhou} \\ 
\midrule
noprompt & 91.5 & 125.2   \\ 
prompt & 70.2 & 70.1 \\
LID & 54.6 & 61.7\\
\bottomrule
\end{tabular}
\caption{Inverse inference \%WERs of Whisper large-v2 on RASC863 dialectal word dataset with different inverse decoding options.}
\label{t3}
\end{table}

\begin{table}[h]
\setlength{\tabcolsep}{3pt}
\centering
\begin{tabular}{c|cccc}
\toprule
\multirow{2}{*}{Setting} & \multicolumn{4}{c}{In-context example number $k$} \\ [0.05cm]
 & $k=1$ & $k=2$ & $k=3$ & $k=4$ \\
\midrule
KATE+ & 67.1 & 54.7 & 51.3 & 49.7 \\ 
$\text{ByCS}_\text{largev2}$ & \textbf{62.4} & 53.4 & 50.6 & \textbf{48.6} \\ 
$\text{ByCS}_\text{small}$ & 64.2 & \textbf{53.3} & \textbf{50.5} & 48.7 \\
\bottomrule
\end{tabular}

\vspace{0.15cm}
(a) Results with Whisper large-v2
\vspace{0.15cm}

\begin{tabular}{c|cccc}
\toprule
\multirow{2}{*}{Setting} & \multicolumn{4}{c}{In-context example number $k$} \\ [0.05cm]
 & $k=1$ & $k=2$ & $k=3$ & $k=4$ \\
\midrule
KATE+ & 68.1 & 58.2 & 54.8 & 54.1 \\ 
$\text{ByCS}_\text{largev3}$ & \textbf{63.5} & \textbf{56.3} & \textbf{53.5} & 51.8 \\ 
$\text{ByCS}_\text{small}$ & 64.4 & 56.5 & 54.1 & \textbf{51.7} \\
\bottomrule
\end{tabular}

\vspace{0.15cm}
(b) Results with Whisper large-v3
\vspace{-0.15cm}

\caption{\%WERs on RASC863 Chongqing dialectal word dataset with ByCS with different inverse inference models. $\text{ByCS}_\text{largev3}$ and $\text{ByCS}_\text{small}$ use Whisper-large-v3 and Whisper-small as the inverse inference model separately.}
\label{t4}
\end{table}

\subsubsection{Text similarity measurement}

The results of ByCS with different text similarity measurements are also reported in Table \ref{t2}. For the setting notation, the ``Jaccard coefficient'' is a commonly used statistic to gauge similarity, defined as the intersection over the union of two sentences. ``BERT wordvecs'' is to measure similarity based on the Euclidean distance in the embedding space of BERT encoded word vectors. The embedding retrieval module is \texttt{bert-base-chinese} \footnote{ \url{https://huggingface.co/bert-base-chinese}}.


ByCS with the Jaccard coefficient as text similarity have lower WERs, which may be because the training data of the BERT model doesn't include sufficient dialectal Chinese words and expressions. 
It also indicates that ByCS can work well with even a simple rule-based text similarity measurement, further verifying its high robustness. 
The Jaccard coefficient is used as the text similarity measurement in later experiments unless explicitly specified, due to the performance and simplicity.

\begin{table*}[t]
\setlength{\tabcolsep}{3pt}
\centering
\begin{tabular}{c|ccc|cc|c|p{1.5cm}<{\centering}p{1.5cm}<{\centering}p{1.5cm}<{\centering}}
\toprule
 & \multicolumn{9}{c}{Corpus \& In-context example number $k$} \\ [0.05cm]
Setting & \multicolumn{3}{c|}{TREC(\%Acc. $\uparrow$)} & \multicolumn{2}{c|}{SST2(\%Acc. $\uparrow$)} & Spider(\%Acc. $\uparrow$) & \multicolumn{3}{c}{HyPoradise CHiME-4}(\%WER $\downarrow$) \\ [0.05cm]
 & $k=1$ & $k=2$ & $k=4$ & $k=1$ & $k=2$ & $k=1$ & $k=1$ & $k=2$ & $k=5$ \\
\midrule
default & \multicolumn{3}{c|}{63.0} & \multicolumn{2}{c|}{92.92} & 67.41 & & 8.0 & \\
random & 63.5 & 72.7 & 75.3 & 94.96 & 94.80 & 67.02 & 7.5 & 7.5 & 7.3\\ 
KATE+ & 78.8 & 86.4 & \textbf{91.0} & 95.05 & 94.69 & 69.44 & 7.7 & 7.1 & 6.8 \\
bm25 & 74.6 & \textbf{89.4} & 89.8 & \textbf{95.27} & \textbf{95.40} & 67.41 & 7.4 & 7.5 & 8.1 \\
LLM-R & 78.0 & 88.8 & 90.4 & 95.05 & 94.02 & 67.82 & 7.4 & 6.9 & 7.0 \\
ByCS & \textbf{81.2} & 88.0 & 90.6 & 95.16 & 95.04 & \textbf{69.63} & \textbf{7.1} & \textbf{6.8} & \textbf{6.4}\\ 
\bottomrule
\end{tabular}

\vspace{0.15cm}
(a) Results using GPT-3.5-Turbo
\vspace{0.15cm}

\begin{tabular}{c|ccc|cc|c|p{1.5cm}<{\centering}p{1.5cm}<{\centering}p{1.5cm}<{\centering}}
\toprule
 & \multicolumn{9}{c}{Corpus \& In-context example number $k$} \\ [0.05cm]
Setting & \multicolumn{3}{c|}{TREC(\%Acc. $\uparrow$)} & \multicolumn{2}{c|}{SST2(\%Acc. $\uparrow$)} & Spider(\%Acc. $\uparrow$) & \multicolumn{3}{c}{HyPoradise CHiME-4 (\%WER $\downarrow$)} \\ [0.05cm]
 & $k=1$ & $k=2$ & $k=4$ & $k=1$ & $k=2$ & $k=1$ & $k=1$ & $k=2$ & $k=5$ \\
\midrule
default & \multicolumn{3}{c|}{75.2} & \multicolumn{2}{c|}{95.01} & 69.63 & & 11.6 &  \\
random & 81.3 & 82.5 & 84.6 & 96.38 & 96.11 & 70.66 & 6.9 & 6.8 & 6.5 \\ 
KATE+ & 88.2 & 91.6 & 93.4 & 96.43 & 95.85 & 71.95 & 7.0 & \textbf{6.3} & \textbf{5.8} \\
bm25 & 81.8 & 87.4 & 91.4 & 96.19 & 96.09 & 71.47 & 6.8 & 6.6 & 6.3 \\
LLM-R & 88.2 & 91.0 & \textbf{93.6} & 95.74 & 95.06 & 72.63 & 6.8 & \textbf{6.3} & 5.9 \\
ByCS & \textbf{88.6} & \textbf{92.4} & \textbf{93.6} & \textbf{96.55} & \textbf{96.31} & \textbf{72.82} & \textbf{6.7} & \textbf{6.3} & 5.9 \\ 
\bottomrule
\end{tabular}

\vspace{0.15cm}
(b) Results using GPT-4
\vspace{-0.15cm}

\caption{Results of four text ICL tasks on two GPT-family models with different in-context example selection methods. The evaluation metrics are denoted in the brackets. The example datastore is narrowed down to a small size using $k$NN for ByCS. In the `default' setting, the answers are generated directly with the questions without ICL.}
\label{t5}
\end{table*}


\subsubsection{Inverse inference model}
\label{sssec:inverse}
The inverse inference with different models is also investigated, with the results displayed in Table \ref{t4}. A smaller model is used for inverse inference to speed up ByCS, since it is expensive to perform inverse inference using the inference model for every candidate example in datastore. Replacing Whisper-large-v2/v3 with Whisper-small will speed up six times\footnote{\url{https://github.com/openai/whisper}}. For the notation, the subscript denotes the inverse inference model. For example, $\text{ByCS}_\text{small}$ is the ByCS method with Whisper small as an inverse inference model.

$\text{ByCS}_\text{small}$ has similar results to $\text{ByCS}_\text{largev2}$ and $\text{ByCS}_\text{largev3}$, verifying the effectiveness of using a smaller model from the same family for inverse inference. This is intuitive since Whisper-small is trained using the same data and settings compared to the inference model Whisper-large-v2 and Whisper-large-v3, which therefore processes information similarly and can serve as a good alternative when evaluating the quality of the in-context examples. The smaller size of Whisper-small makes ByCS a more practical method in cost-sensitive scenarios. A detailed analysis of time cost is in Appendix \ref{sec:appendixb}.




\subsection{Text ICL}

\begin{figure}[h]
\centering

\vspace{-0.1cm}
\includegraphics[width=0.42\textwidth,height=0.28\textwidth]{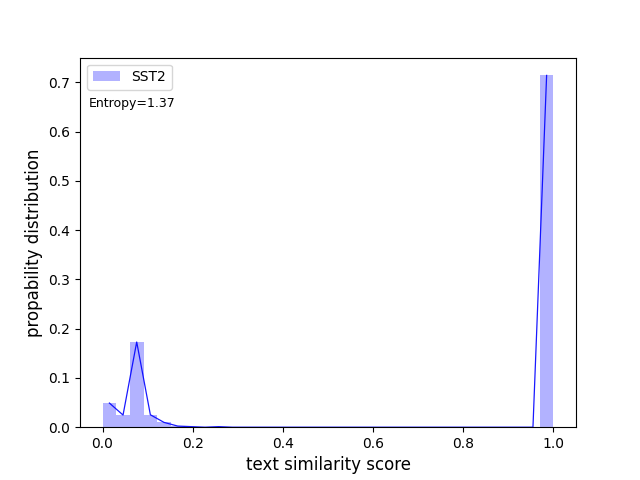}

\vspace{0.05cm}
{\fontsize{9.5}{\baselineskip}\selectfont (a) Distribution on SST2}
\vspace{-0.04cm}

\includegraphics[width=0.42\textwidth,height=0.28\textwidth]{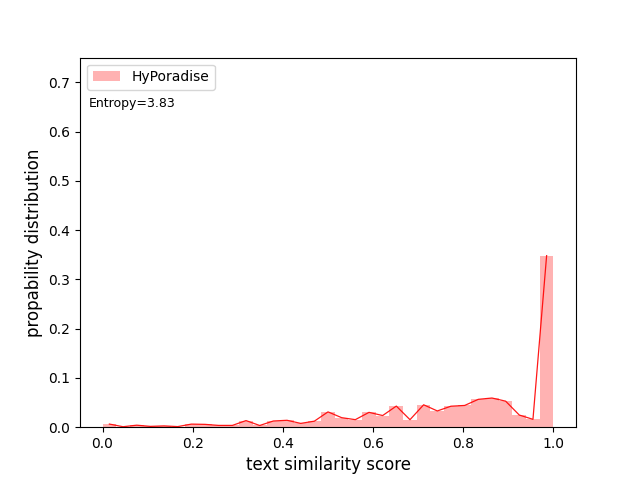}

\vspace{0.05cm}
{\fontsize{9.5}{\baselineskip}\selectfont (b) Distribution on HyPoradise}
\vspace{-0.1cm}

\caption{The distribution of text similarity scores on different datasets. The text similarity score is the Jaccard coefficient. The entropy of distribution is calculated and placed on the upper left. The distribution on the multichoice classification dataset SST2 (blue) is much sharper than that of the open-ended dataset HyPoradise (red).}
\label{f4}
\end{figure}

Text-only ICL results are shown in Table \ref{t5}. As shown, ByCS outperforms all baselines on most dataset settings, showing not only the effectiveness but also the robustness of ByCS. In particular, ByCS outperforms the best baseline on the generative ASR rescoring dataset HyPoradise with a considerable 4.7\% relative WER reduction with GPT-3.5-Turbo. On TREC and SST2 datasets, ByCS does not always outperform the baselines. This indicates that ByCS is more suitable for open-ended long-answer datasets due to the calculation of text similarity in ByCS, in which answers are much more diverse and examples with rich information interactions can be better separated. In contrast, in multi-choice classification datasets, only a few short answers are often available, containing little contextual information. As the example shown in Figure \ref{f4}, the distribution of the text similarity for ranking the examples is often sharp, merging the optimal and the suboptimal examples. Furthermore, considering the hypothesized labels of the test inputs for inverse inference, the hypothesized answers in open-ended datasets (in the form of long sentences) are often more similar to their corresponding references compared to those in the multi-choice classification datasets (in the form of a word or phrase or just an index of choice).

It is observed that different in-context example selection methods perform differently with different models, even though on the same dataset. The bm25 method outperforms the KATE+ method with GPT-3.5-Turbo on the SST2 dataset, but not with GPT4. Compared to KATE+ and bm25 that is model-free in the actual selection step, the performance advantage of ByCS is more consistent since it takes into account the influence of the model. The outputs of the inverse inference model are used, which can serve as a good approximation to the inference model as verified in Section~\ref{sssec:inverse}.

Note that for ByCS on GPT-4, although the inverse inference procedure is conducted on GPT-3.5-Turbo, the performances of ByCS are still superior. This further verifies that smaller models from the same model family can serve as a good low-cost approximation of the inverse inference model.


\vspace{-0.1cm}

\subsection{VQA ICL}

\begin{table}[h]
\vspace{0.1cm}
\setlength{\tabcolsep}{3pt}
\centering
\begin{tabular}{c|p{2cm}<{\centering}p{2cm}<{\centering}}
\toprule
\multirow{2}{*}{\makecell{In-context \\ example \\ number $k$}} & \multicolumn{2}{c}{Example selection method} \\ [0.05cm]
 & \multirow{2}{*}{KATE+} & \multirow{2}{*}{ByCS} \\ 
 & & \\
\midrule
k = 2 & \textbf{40.47} & 40.12   \\ 
k = 4 & 45.11 & \textbf{45.14} \\
\bottomrule
\end{tabular}

\vspace{0.15cm}
(a) Results with Emu-2
\vspace{0.15cm}

\begin{tabular}{c|p{2cm}<{\centering}p{2cm}<{\centering}}
\toprule
\multirow{2}{*}{\makecell{In-context \\ example \\ number $k$}} & \multicolumn{2}{c}{Example selection method} \\ [0.05cm]
 & \multirow{2}{*}{KATE+} & \multirow{2}{*}{ByCS}  \\ 
 & & \\
\midrule
k = 2 & 52.54 & \textbf{52.86}   \\ 
k = 4 & 54.00 & \textbf{54.39} \\
\bottomrule
\end{tabular}

\vspace{0.15cm}
(b) Results with GPT-4V
\vspace{-0.15cm}
\caption{Results of VQA ICL with different in-context example selection methods and numbers of examples on OKVQA dataset.}
\label{t6}
\end{table}


ByCS is tested on VQA ICL and the results are reported in Table \ref{t6}. ByCS outperforms the KATE+ baseline on the VQA ICL task, demonstrating strong performances across modalities. The performance improvement from ByCS is not as obvious as in audio and text tasks, since the answers of VQA are usually short (usually a word or phrase), lacking sufficient contextual information. ByCS on the VQA dataset suffers from the problem of having sharp text similarity score distributions, similar to the multichoice classification dataset. For ByCS with GPT-4V, inverse inference results on Emu-2 are used to pre-select the candidate examples, and ByCS still outperforms the KATE+ baseline. The performance may be further improved if GPT-4V is also used for inverse inference. This demonstrates that ICL may perform similarly cross models not only on speech and text, but also on images.


\section{Conclusion}

This paper proposes ByCS, a novel in-context example selection method based on Bayes' theorem, which assumes that contextual information interaction is mutual between the test input and in-context examples and selects high-quality examples based on the inverse inference results.  
Experiments are performed across three modalities: speech, text, and images, using six different tasks and seven datasets. Results demonstrated the robustness and effectiveness of ByCS. It is also validated that the inverse inference results can be approximated using a smaller model from the same model family, which considerably reduces the computational cost. Moreover, relying on text similarity to rank in-context examples, ByCS is more suitable for open-ended long-answer datasets which contain sufficient contextual information. Future work is to extend the inverse inference to sequences with multiple in-context examples to model the interactions among the in-context examples.

\section*{Limitations}
There are three limitations to this work.
First, ByCS follows the simple assumption that the influence of each in-context example is independent and treats each in-context example individually, which neglects the contextual interactions between in-context examples. The approximation may not be adapted to the scenario in which the number of in-context examples is high.
Second, ByCS requires sufficient contextual diversity to select optimal examples, which depends on text similarity to evaluate inverse inference results. ByCS may suffer a performance penalty when applied to a short-answer dataset. The third limitation is the extra time cost introduced by inverse inference, making ByCS less suitable for cost-sensitive scenarios. Future work includes enhancing ByCS in more scenarios.

\section*{Ethics Statement}
The work doesn't give rise to any ethical risks and issues. All the models and data used in this paper are publicly accessible and used under licenses.

\bibliography{anthology,custom}
\bibliographystyle{acl_natbib}

\appendix
\clearpage

\section{Experimental Details}
\label{sec:appendix}
\subsection{Datasets, baselines and prompt templates}

\begin{figure}[h]
\centering
\includegraphics[width=3.1in]{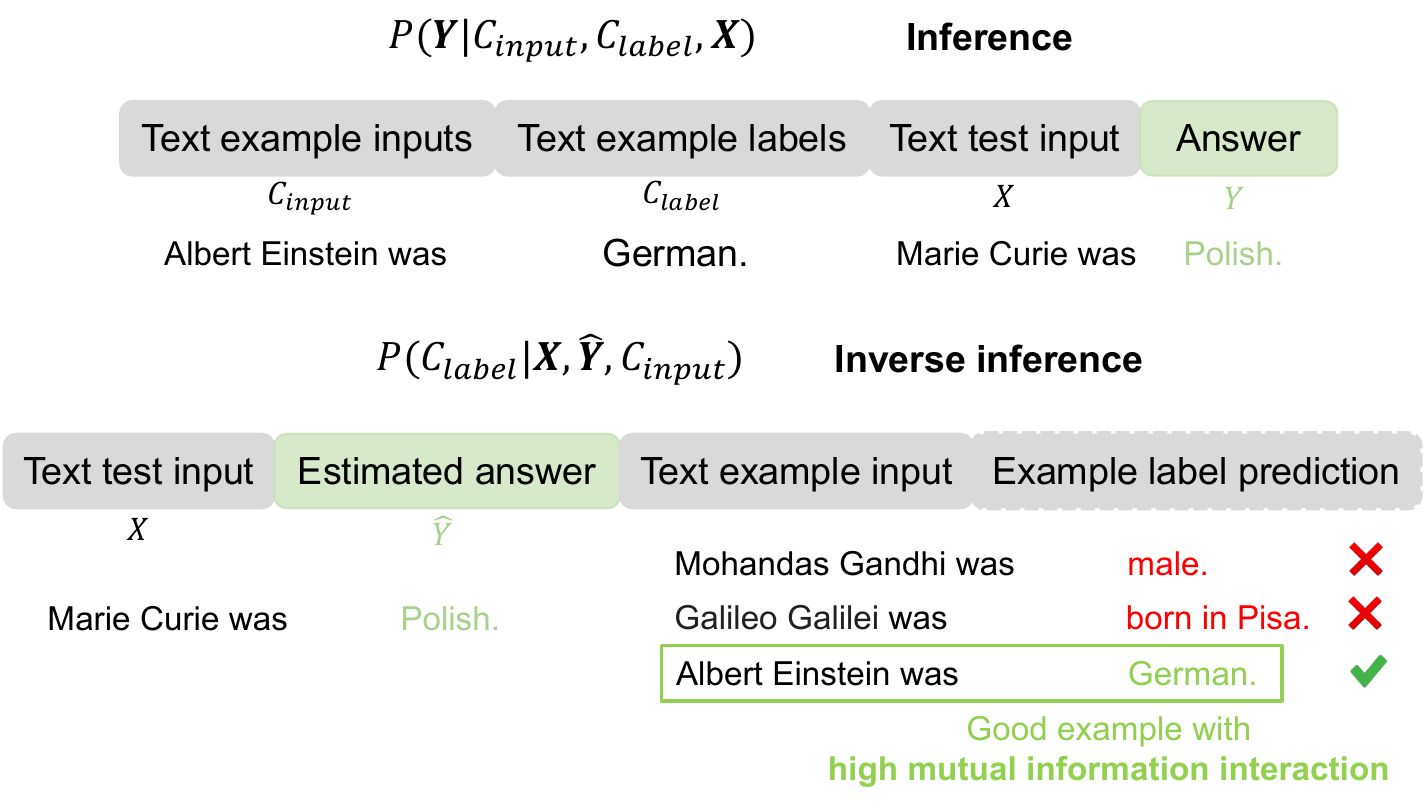}
\vspace{-0.4cm}
\caption{We provide an additional ``inverse inference'' illustration of the proposed Bayesian example selection method for in-context learning in a text format, similar to \citet{min2022rethinking}.}
\label{fig:5}
\end{figure}

The dataset details are listed in Table \ref{t7}. For Spider, the evaluation metric is execution accuracy. For CORAAL, we use the processing script from the FairSpeech project\footnote{\url{https://github.com/stanford-policylab/asr-disparities}}. For convenience, we only use speech less than 15 seconds because Whisper can accept input audio up to 30 seconds. For the ASR dataset, there is no train/test split, the dataset except the test input serves as the in-context example datastore. For bm25 implementation, we use the okapi variant in $\text{rank\_bm25}\footnote{\url{https://github.com/dorianbrown/rank_bm25}}$ library. The inverse inference example is presented in Figure~\ref{fig:5} and prompt templates are shown in Table \ref{t8}.

\begin{figure}[h]
\centering
\includegraphics[width=2.1in]{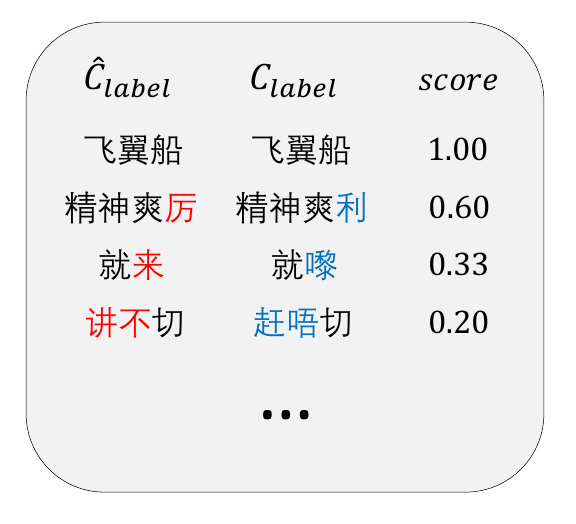}
\vspace{-0.4cm}
\caption{An illustration of the calculation of text similarity between inverse inference results and their true labels in Mandarin accent recognition, where the red inverse inference tokens indicate misrecognition.}
\label{fig:6}
\end{figure}

\subsection{First-round inference of ByCS}

We experimented with ByCS on different first-round inference settings to examine the influence of first-round inference, and the results are reported in Table \ref{firstround}. The first-round inference produces the hypothesized label of test input. With better first-round inference hypotheses, the approximated inverse inference probability will be more close to the oracle one. Figure~\ref{fig:6} provides an example of text similarity calculation. The first-round accuracy for the `default', `random' and `KATE+' settings is 63.0, 75.8 and 91.0, respectively. The first-round inference with ICL improves the accuracy of the hypothesized label, thus boosting the performance of ByCS. In practice, we use ICL with random example selection as the first-round inference setting for ASR ICL and best ICL baseline as the first-round inference setting for text and VQA ICL.

\begin{table}[h]
\vspace{0.1cm}
\setlength{\tabcolsep}{3pt}
\centering
\resizebox{0.48\textwidth}{!}{
\begin{tabular}{c|p{1.35cm}<{\centering}p{1.35cm}<{\centering}p{1.35cm}<{\centering}}
\toprule
\multirow{2}{*}{\makecell{First-round \\ inference}} & \multicolumn{3}{c}{In-context example number $k$} \\ [0.05cm]
 & $k=1$ & $k=2$ & $k=4$ \\
\midrule
default & 75.6 & 83.8 & 88.4 \\ 
random $k=4$ & 79.8 & 87.0 & 91.6 \\ 
KATE+ $k=4$ & 81.2 & 88.0 & 90.6  \\ 
\bottomrule
\end{tabular}
}

\vspace{0.15cm}
(a) Results with GPT-3.5-Turbo
\vspace{0.15cm}

\resizebox{0.48\textwidth}{!}{
\begin{tabular}{c|p{1.35cm}<{\centering}p{1.35cm}<{\centering}p{1.35cm}<{\centering}}
\toprule
\multirow{2}{*}{\makecell{First-round \\ inference}} & \multicolumn{3}{c}{In-context example number $k$} \\ [0.05cm]
 & $k=1$ & $k=2$ & $k=4$ \\
\midrule
default & 87.2 & 91.8 & 93.0 \\ 
random $k=4$ & 86.6 & 92.4 & 93.0 \\ 
KATE+ $k=4$ & 88.6 & 92.4 & 93.6  \\ 
\bottomrule
\end{tabular}
}

\vspace{0.15cm}
(b) Results with GPT-4
\vspace{-0.15cm}

\caption{Results on TREC of ByCS with different first-round inference settings.}
\label{firstround}
\end{table}

\subsection{Pre-selection of ByCS}
\label{sec:appendixa3}

Since the datastore size is usually large, we use a simple ranking algorithm to compress in-context example datastore and then use ByCS inverse inference to select good examples. We usually choose $k$NN as the ranking algorithm and twice the maximum number of in-context examples as reduced size after pre-selection. For RASC863, we simply use the speech from the same speaker as in-context examples, so the number of reduced size is approximate. We experimented on the TREC dataset to analyze whether reduced size matters, the results are reported in \ref{tt}. The results imply that reduced size has nearly negligible impact on the performance of ByCS method. Thus twice the number of in-context examples is a balanced choice for example diversity and conducting speed. The details of pre-selection are shown in Table \ref{t7}.

\begin{table}[h]
\vspace{0.1cm}
\setlength{\tabcolsep}{3pt}
\centering
\begin{tabular}{c|p{1.35cm}<{\centering}p{1.35cm}<{\centering}p{1.35cm}<{\centering}}
\toprule
\multirow{2}{*}{\makecell{Reduced \\ size}} & \multicolumn{3}{c}{In-context example number $k$} \\ [0.05cm]
 & $k=1$ & $k=2$ & $k=4$ \\
\midrule
 4 & 81.6 & 87.6 & 90.6 \\ 
 8 & 81.2 & 88.0 & 90.6  \\ 
 16 & 81.0 & 88.0 & 90.4 \\ 
\bottomrule
\end{tabular}

\vspace{0.15cm}
(a) results on GPT-3.5-Turbo
\vspace{0.15cm}

\begin{tabular}{c|p{1.35cm}<{\centering}p{1.35cm}<{\centering}p{1.35cm}<{\centering}}
\toprule
\multirow{2}{*}{\makecell{Reduced \\ size}} & \multicolumn{3}{c}{In-context example number $k$} \\ [0.05cm]
& $k=1$ & $k=2$ & $k=4$ \\
\midrule
4 & 88.0 & 92.6 & 93.2 \\ 
8 & 88.6 & 92.4 & 93.6  \\ 
16 & 88.4 & 92.8 & 93.2 \\ 
\bottomrule
\end{tabular}

\vspace{0.15cm}
(b) results on GPT-4
\vspace{-0.15cm}

\caption{Results on TREC of ByCS with different reduced sizes after pre-selection.}
\label{tt}
\end{table}

\begin{table*}[h]
\vspace{0.1cm}
\setlength{\tabcolsep}{3pt}
\centering
\resizebox{\textwidth}{!}{
\begin{tabular}{ccccccc}
\toprule
Modality & Task category & Dataset & Train size & Test size & \makecell{Pre-selection} & \makecell{Reduced \\ size} \\ [0.05cm]
\midrule
\multirow{4}{*}{Text} & Topic classification & TREC & 5452 & 500 & \makecell{$k$NN} & 8 \\ 
 & Sentiment analysis & SST2 & 67349 & 872 & \makecell{$k$NN} & 4 \\ 
 & Text to SQL & Spider & 7000 & 1034 & $k$NN & 3\\ 
 & ASR LM rescoring & HyPoradise CHiME-4 & 9728 & 1320 & $k$NN & 10 \\ 
\hline
\multirow{3}{*}{Audio} & \multirow{3}{*}{Automatic speech recognition} & RASC863 Guangzhou & 1889 & 1990(1.41h) & \makecell{same  speaker} & $\sim$ 10 \\
 & & RASC863 Chongqing & 2993 & 2994(3.26h) & \makecell{same  speaker} & $\sim$ 15 \\
 & & CORAAL <15s & 2761 & 2762(6.77h) & $k$NN & 10 \\
\hline
Image & Vision question answering & OKVQA & 9009 & 5046 & $k$NN & 8\\ 
\bottomrule
\end{tabular}
}
\caption{Datasets used in this work}
\label{t7}
\end{table*}

\begin{table*}[h]
\vspace{0.1cm}
\setlength{\tabcolsep}{3pt}
\centering
\resizebox{\textwidth}{!}{%
\begin{tabular}{cc}
\toprule
Dataset & Template example \\ [0.05cm]
\midrule
TREC & \makecell[l]{Question: \textcolor{blue}{What is the temperature at the centre of the earth?} \\ Available Type: description, entity, expression, human, number, location. \\ Type: \textcolor{blue}{number}.} \\ 
\hline
SST2 & \makecell[l]{Review: \textcolor{blue}{``The Time Machine'' is a movie that has no interest in itself.} \\ Available sentiment: positive, negative. \\ Sentiment: \textcolor{blue}{negative}.} \\ 
\hline
Spider & \makecell[l]{Given the database schema, you need to translate the question into the SQL query. \\ Database schema: \\
\textcolor{blue}{Table name: Movie} \\
\textcolor{blue}{Creation SQL: CREATE TABLE Movie(} \\
\textcolor{blue}{	mID int primary key,} \\
\textcolor{blue}{	title text,} \\
\textcolor{blue}{	year int,} \\
\textcolor{blue}{	director text} \\
\textcolor{blue}{)} \\
\textcolor{blue}{Table name: Reviewer} \\
\textcolor{blue}{Creation SQL: CREATE TABLE Reviewer(} \\
\textcolor{blue}{	rID int primary key,} \\
\textcolor{blue}{	name text} \\
\textcolor{blue}{)} \\
\textcolor{blue}{Table name: Rating} \\
\textcolor{blue}{Creation SQL: CREATE TABLE Rating(} \\
\textcolor{blue}{	rID int,} \\
\textcolor{blue}{	mID int,} \\
\textcolor{blue}{	stars int,} \\
\textcolor{blue}{	ratingDate date,} \\
\textcolor{blue}{	FOREIGN KEY (mID) references Movie(mID),} \\
\textcolor{blue}{	FOREIGN KEY (rID) references Reviewer(rID)} \\
\textcolor{blue}{)} \\ Question: \textcolor{blue}{Find the names of all reviewers who have contributed three or more ratings.} \\ SQL query: \textcolor{blue}{SELECT T2.name FROM Rating AS T1 JOIN Reviewer AS T2 ON T1.rID = T2.rID GROUP BY T1.rID HAVING COUNT(*) >= 3.}} \\ 
\hline
\makecell{HyPoradise \\ CHiME-4} & \makecell[l]{You need to do language model rescoring in ASR. Given the 5-best hypotheses, you need to report the true transcription from the 5-best hypotheses.\\ The 5-best hypothesis is: \\ \textcolor{blue}{interest rates rose on torture and treasury bills sold by the government yesterday at its regular weekly auction.} \\ \textcolor{blue}{interest rates rose on short-term treasury bills sold by the government yesterday at its regular weekly auction.}  \\ \textcolor{blue}{interest rates rose at a torture and treasury bill sold by the government yesterday at its regular weekly auction.}  \\ \textcolor{blue}{interest rates rose on a torture and treasury bill sold by the government yesterday at its regular weekly auction.}  \\ \textcolor{blue}{interest rates rose on torturing treasury bills sold by the government yesterday at its regular weekly auction.}\\ The true transcription from the 5-best hypotheses is: \\ \textcolor{blue}{interest rates rose on short-term treasury bills sold by the government yesterday at its regular weekly auction.}}  \\ 
\hline
OKVQA & \makecell[l]{\includegraphics[width=1.6in]{example_okvqa.jpg} \\ Answer in one word or phrase. \\ \textcolor{blue}{What softwood is used to close the top of the container in his hand?} \\ \textcolor{blue}{cork.}}  \\ 
\bottomrule
\end{tabular}%
}
\caption{Prompt template examples used in this work}
\label{t8}
\end{table*}

\begin{table*}[t]
\setlength{\tabcolsep}{3pt}
\centering
\begin{tabular}{cc|p{1.33cm}<{\centering}p{1.33cm}<{\centering}p{1.33cm}<{\centering}p{1.33cm}<{\centering}p{1.33cm}<{\centering}p{1.33cm}<{\centering}}
\toprule
\multirow{3}{*}{\makecell{In-context \\ example \\ number $k$}} & \multirow{3}{*}{\makecell{Inverse \\ inference \\ model}} & \multicolumn{6}{c}{Text similarity measurement \& inverse decoding option} \\ [0.05cm]
 & & \multicolumn{3}{c}{Jaccard coefficient} & \multicolumn{3}{c}{BERT wordvecs}  \\ [0.05cm]
  & & noprompt & prompt & LID & noprompt & prompt & LID\\ [0.05cm]
\midrule
\multirow{2}{*}{$k=1$} & $\text{ByCS}_{largev2}$ & \textbf{62.4} & 62.9 & 64.1 & \textbf{62.4} & 63.5 & 64.5\\ 
&$\text{ByCS}_{small}$ & 64.2 & 64.0 & 65.4 & 65.0 & 65.4 & 66.3 \\ 
\multirow{2}{*}{$k=2$} & $\text{ByCS}_{largev2}$ & 53.4 & \textbf{53.3} & 53.7 & 53.6 & 54.1 & 54.1 \\ 
&$\text{ByCS}_{small}$ & \textbf{53.3} & 53.7 & 54.0 & 54.1 & 54.9 & 54.8 \\
\multirow{2}{*}{$k=3$} & $\text{ByCS}_{largev2}$ & 50.6 & 51.0 & 50.9 & \textbf{50.2} & 51.6 & 50.6 \\ 
&$\text{ByCS}_{small}$ & 50.5 & 50.5 & 51.1 & 51.3 & 50.9 & 51.3 \\
\multirow{2}{*}{$k=4$} & $\text{ByCS}_{largev2}$ & \textbf{48.6} & 48.7 & 48.7 & 49.1 & 48.9 & 49.1 \\ 
&$\text{ByCS}_{small}$ & 48.7 & 48.7 & \textbf{48.6} & 49.6 & 49.1 & 49.9 \\
\bottomrule
\end{tabular}

\vspace{0.3cm}
(a) Results with Whisper large-v2
\vspace{0.25cm}

\begin{tabular}{cc|p{1.33cm}<{\centering}p{1.33cm}<{\centering}p{1.33cm}<{\centering}p{1.33cm}<{\centering}p{1.33cm}<{\centering}p{1.33cm}<{\centering}}
\toprule
\multirow{3}{*}{\makecell{In-context \\ example \\ number $k$}} & \multirow{3}{*}{\makecell{Inverse \\ inference \\ model}} & \multicolumn{6}{c}{Text similarity measurement \& inverse decoding option} \\ [0.05cm]
 & & \multicolumn{3}{c}{Jaccard coefficient} & \multicolumn{3}{c}{BERT wordvecs}  \\ [0.05cm]
  & & noprompt & prompt & LID & noprompt & prompt & LID\\ [0.05cm]
\midrule
\multirow{2}{*}{$k=1$} & $\text{ByCS}_{largev3}$ & \textbf{63.5} & 64.1 & 65.6 & 64.5 & 65.3 & 65.8\\ 
&$\text{ByCS}_{small}$ & 64.4 & 64.7 & 64.8 & 65.5 & 65.0 & 65.6  \\ 
\multirow{2}{*}{$k=2$} & $\text{ByCS}_{largev3}$ & \textbf{56.3} & \textbf{56.3} & 57.0 & 57.7 & 57.0 & 57.8 \\ 
&$\text{ByCS}_{small}$ & 56.5 & 57.0 & 57.0 & 57.3 & 57.2 & 57.5 \\
\multirow{2}{*}{$k=3$} & $\text{ByCS}_{largev3}$ & \textbf{53.5} & 54.1 & 53.7 & 55.2 & 55.6 & 54.9 \\ 
&$\text{ByCS}_{small}$ & 54.1 & 54.6 & 54.4 & 55.5 & 55.3 & 55.4 \\
\multirow{2}{*}{$k=4$} & $\text{ByCS}_{largev3}$ & 51.8 & 52.3 & 52.1 & 53.1 & 53.4 & 53.3\\ 
&$\text{ByCS}_{small}$ & \textbf{51.7} & 52.2 & 51.9 & 53.6 & 53.4 & 53.5 \\
\bottomrule
\end{tabular}

\vspace{0.3cm}
(b) Results with Whisper large-v3
\vspace{-0.15cm}

\caption{Full results on RASC863 Chongqing dialectal word dataset of ByCS with different inverse decoding options, text similarity measurements and inverse inference models. The subscript denotes the inverse inference model.}
\end{table*}

\begin{table*}[t]
\setlength{\tabcolsep}{3pt}
\centering
\begin{tabular}{cc|p{1.33cm}<{\centering}p{1.33cm}<{\centering}p{1.33cm}<{\centering}p{1.33cm}<{\centering}p{1.33cm}<{\centering}p{1.33cm}<{\centering}}
\toprule
\multirow{3}{*}{\makecell{In-context \\ example \\ number $k$}} & \multirow{3}{*}{\makecell{Inverse \\ inference \\ model}} & \multicolumn{6}{c}{Text similarity measurement \& inverse decoding option} \\ [0.05cm]
 & & \multicolumn{3}{c}{Jaccard coefficient} & \multicolumn{3}{c}{BERT wordvecs}  \\ [0.05cm]
  & & noprompt & prompt & LID & noprompt & prompt & LID\\ [0.05cm]
\midrule
\multirow{2}{*}{$k=1$} & $\text{ByCS}_\text{largev2}$ & \textbf{49.5} & 50.7 & 52.3 & 51.5 & 56.8 & 57.7\\ 
&$\text{ByCS}_\text{small}$ & 52.9 & 55.1 & 58.7 & 56.8 & 57.1 & 58.8 \\ 
\multirow{2}{*}{$k=2$} & $\text{ByCS}_\text{largev2}$ & \textbf{31.9} & 33.6 & 34.3 & 32.9 & 34.3 & 35.0 \\ 
&$\text{ByCS}_\text{small}$ & 34.5 & 34.1 & 35.6 & 35.1 & 35.9 & 37.0 \\
\multirow{2}{*}{$k=3$} & $\text{ByCS}_\text{largev2}$ & \textbf{27.1} & 28.4 & 27.7 & \textbf{27.1} & 27.4 & 27.5 \\ 
&$\text{ByCS}_\text{small}$ & 28.3 & 27.8 & 27.6 & 27.9 & 28.6 & 28.3 \\
\multirow{2}{*}{$k=4$} & $\text{ByCS}_\text{largev2}$ & 26.6 & 25.5 & \textbf{24.8} & 25.4 & 26.5 & 25.5\\ 
&$\text{ByCS}_\text{small}$ & 25.9 & 25.7 & 25.5 & 25.3 & 26.3 & 26.2 \\
\bottomrule
\end{tabular}

\vspace{0.3cm}
(a) Results with Whisper large-v2
\vspace{0.25cm}

\begin{tabular}{cc|p{1.33cm}<{\centering}p{1.33cm}<{\centering}p{1.33cm}<{\centering}p{1.33cm}<{\centering}p{1.33cm}<{\centering}p{1.33cm}<{\centering}}
\toprule
\multirow{3}{*}{\makecell{In-context \\ example \\ number $k$}} & \multirow{3}{*}{\makecell{Inverse \\ inference \\ model}} & \multicolumn{6}{c}{Text similarity measurement \& inverse decoding option} \\ [0.05cm]
 & & \multicolumn{3}{c}{Jaccard coefficient} & \multicolumn{3}{c}{BERT wordvecs}  \\ [0.05cm]
  & & noprompt & prompt & LID & noprompt & prompt & LID\\ [0.05cm]
\midrule
\multirow{2}{*}{$k=1$} & $\text{ByCS}_\text{largev3}$ & \textbf{50.7} & 51.8 & 55.4 & 56.6 & 57.1 & 59.1\\ 
&$\text{ByCS}_\text{small}$ & 55.3 & 55.4 & 61.7 & 61.8 & 58.7 & 60.7  \\ 
\multirow{2}{*}{$k=2$} & $\text{ByCS}_\text{largev3}$ & \textbf{36.7} & 38.1 & 38.9 & 38.2 & 37.8 & 38.9 \\ 
&$\text{ByCS}_\text{small}$ & 37.3 & 37.3 & 40.0 & 39.0 & 38.0 & 39.6 \\
\multirow{2}{*}{$k=3$} & $\text{ByCS}_\text{largev3}$ & \textbf{33.0} & 33.4 & 34.0 & 33.6 & 33.4 & 33.3 \\ 
&$\text{ByCS}_\text{small}$ & 33.3 & 33.3 & 34.6 & 34.8 & 33.3 & 34.3 \\
\multirow{2}{*}{$k=4$} & $\text{ByCS}_\text{largev3}$ & 31.5 & 31.3 & 31.4 & 31.7 & 31.7 & 31.4\\ 
&$\text{ByCS}_\text{small}$ & \textbf{31.0} & 31.5 & 31.9 & 31.5 & \textbf{31.0} & \textbf{31.0} \\
\bottomrule
\end{tabular}

\vspace{0.3cm}
(b) Results with Whisper large-v3
\vspace{-0.15cm}

\caption{Full results on RASC863 Guangzhou dialectal word dataset of ByCS with different inverse decoding options, text similarity measurements and inverse inference models. The subscript denotes the inverse inference model.}
\end{table*}

\begin{table*}[h]
\vspace{0.1cm}
\setlength{\tabcolsep}{3pt}
\centering
\resizebox{\textwidth}{!}{%
\begin{tabular}{ccc}
\toprule
Test input & KATE+ & ByCS \\ [0.05cm]
\midrule
\makecell{sometime they do not act like they hear nothing \\ but know nothing about tarboro \\ when you say you from tarboro \\ they will talk about where is tarboro at \\ (CORAAL)} & \makecell{Example: \\ in the era and th the way \\ in there them floors along that time \\ they cut timber certain time of the year \\ Result: \\sometimes \textcolor{red}{it} do not \textcolor{red}{work out there} \\ but \textcolor{red}{no} nothing about \textcolor{red}{tarver} \\ when you say you from \textcolor{red}{tarver} \\ they will talk about where \textcolor{red}{tarver is}} & \makecell{Example: \\ so they put her and him together \\ and i was praying to the lord \\ that he did not try to jump out of there \\ cause i was so scared me and my husband \\Result: \\ sometimes they do not \textcolor{red}{want to let their} hear nothing\\ but know nothing about \textcolor{red}{tarver} \\ when you say you from \textcolor{red}{tarver} \\they will talk about where \textcolor{red}{tarver is}} \\
\hline
\makecell{What person 's head is on a dime? \\ \textcolor{blue}{human.} \\ (TREC)} & \makecell{Example: \\ What is money made of? \\ \textcolor{blue}{entity.} \\ Result: \\ \textcolor{red}{entity.}} & \makecell{Example: \\ Who is the head of the World Bank? \\ \textcolor{blue}{human.} \\ Result: \\ \textcolor{blue}{human.}} \\
\bottomrule
\end{tabular}%
}
\caption{In-context examples selected by $k$NN and ByCS and corresponding results.}
\label{t8}
\end{table*}

\section{Analysis of time cost}
\label{sec:appendixb}

\subsection{Computational complexity}
Although ByCS may be time-consuming, the existing improvement methods have reduced the complexity from $\mathcal{O}(N)$ to $\mathcal{O}(1)$, where $N$ is the size of the example datastore. The original version of ByCS will conduct inverse inference on every candidate in the whole dataset, which results in complexity in $\mathcal{O}(N)$. Using a smaller model for fast inverse inference decreases the number of computations by a constant factor. For instance, Whisper small is 6 times faster than Whisper large, and using Whisper small for inverse inference reduces the inverse inference cost by $\sim$6 times. Furthermore, by using a ranking-based pre-selection, we can reduce the size of the example datastore to a fixed number, reducing the computational complexity of inverse inference further down to $\mathcal{O}(1)$. In our experiments, we found empirically that a number around 10 is a good choice in balancing the example diversity and conduction speed, as shown in Appendix \ref{sec:appendixa3}.

\subsection{Attempt to further speed up}
Since inverse inference spends most of its time in ByCS, we try to conduct inverse inference on examples in the datastore before the test input arrives. For each example in the datastore, suitable in-context examples are selected for it using ByCS. In practice, the in-context examples of the test input are those of the nearest neighbour. By this means, the time cost of ByCS is comparable with $k$NN-based methods. The results of this new sped-up version of ByCS, which is denoted as $\text{ByCS}_\text{fast}$ are shown in Table \ref{t14}. As expected,  $\text{ByCS}_\text{fast}$ always performs worse than ByCS. Furthermore, $\text{ByCS}_\text{fast}$ is more dependent on the contextual diversity. On the open-ended long-answer speech datasets, $\text{ByCS}_\text{fast}$ can outperform the best baseline. While on short-answer text datasets, the performance of $\text{ByCS}_\text{fast}$ suffers a significant deterioration. It emphasizes the importance of inverse inference directly on test input, not on a similar substitution.

\begin{table*}[h]
\setlength{\tabcolsep}{3pt}
\centering
\begin{tabular}{c|cccc|cccc|c}
\toprule
 & \multicolumn{9}{c}{Corpus \& In-context example number $k$} \\ [0.05cm]
Setting & \multicolumn{4}{c|}{RASC863 Chongqing} & \multicolumn{4}{c|}{RASC863 Guangzhou} & CORAAL <15s\\ [0.05cm]
 & $k=1$ & $k=2$ & $k=3$ & $k=4$ & $k=1$ & $k=2$ & $k=3$ & $k=4$ & $k=1$ \\
\midrule
best baseline & 67.1 & 54.7 & 51.3 & 49.7 & 61.3 & 36.1 & \textbf{26.9} & \textbf{24.8} & 12.6\\ 
$\text{ByCS}_\text{fast}$ & 63.1 & \textbf{52.5} & \textbf{50.2} & \textbf{48.3} & 55.8 & 35.6 & 29.2 & 27.1 & 12.5\\
ByCS & \textbf{62.4} & 53.4 & 50.6 & 48.6 & \textbf{49.5} & \textbf{31.9} & 27.1 & 26.6 & \textbf{12.4}\\ 
\bottomrule
\end{tabular}

\vspace{0.15cm}
(a) Results with Whisper-large-v2
\vspace{0.15cm}

\begin{tabular}{c|cccc|cccc|c}
\toprule
 & \multicolumn{9}{c}{Corpus \& In-context example number $k$} \\ [0.05cm]
Setting & \multicolumn{4}{c|}{RASC863 Chongqing} & \multicolumn{4}{c|}{RASC863 Guangzhou} & CORAAL <15s\\ [0.05cm]
 & $k=1$ & $k=2$ & $k=3$ & $k=4$ & $k=1$ & $k=2$ & $k=3$ & $k=4$ & $k=1$ \\
\midrule
best baseline & 68.1 & 58.2 & 54.8 & 54.1 & 67.1 & 41.3 & 34.3 & 31.6 & 12.1\\ 
$\text{ByCS}_\text{fast}$ & 66.7 & 57.5 & 54.5 & 52.6 & 60.5 & 40.3 & 34.1 & 32.3 & 12.2 \\
ByCS & \textbf{63.5} & \textbf{56.3} & \textbf{53.5} & \textbf{51.8} & \textbf{50.7} & \textbf{36.7} & \textbf{33.0} & \textbf{31.5} & \textbf{12.0}\\ 
\bottomrule
\end{tabular}

\vspace{0.15cm}
(b) Results with Whisper-large-v3
\vspace{0.15cm}

\setlength{\tabcolsep}{3pt}
\centering
\begin{tabular}{c|ccc|cc}
\toprule
 & \multicolumn{5}{c}{Corpus \& In-context example number $k$} \\ [0.05cm]
Setting & \multicolumn{3}{c|}{TREC(\%Acc. $\uparrow$)} & \multicolumn{2}{c}{SST2(\%Acc. $\uparrow$)} \\ [0.05cm]
 & $k=1$ & $k=2$ & $k=4$ & $k=1$ & $k=2$ \\
\midrule
best baseline & 78.8 & \textbf{89.4} & \textbf{91.0} & \textbf{95.27} & \textbf{95.40} \\
$\text{ByCS}_\text{fast}$ & 77.0 & 83.8 & 86.4 & 94.15 & 94.61 \\
ByCS & \textbf{81.2} & 88.0 & 90.6 & 95.16 & 95.04 \\ 
\bottomrule
\end{tabular}

\vspace{0.15cm}
(c) Results using GPT-3.5-Turbo
\vspace{0.15cm}

\begin{tabular}{c|ccc|cc}
\toprule
 & \multicolumn{5}{c}{Corpus \& In-context example number $k$} \\ [0.05cm]
Setting & \multicolumn{3}{c|}{TREC(\%Acc. $\uparrow$)} & \multicolumn{2}{c}{SST2(\%Acc. $\uparrow$)} \\ [0.05cm]
 & $k=1$ & $k=2$ & $k=4$ & $k=1$ & $k=2$ \\
\midrule
best baseline & 88.2 & 91.6 & \textbf{93.6} & 96.43 & 96.11 \\
$\text{ByCS}_\text{fast}$ & 85.4 & 89.2 & 92.6 & 95.07 & 95.18  \\
ByCS & \textbf{88.6} & \textbf{92.4} & \textbf{93.6} & \textbf{96.55} & \textbf{96.31} \\ 
\bottomrule
\end{tabular}

\vspace{0.15cm}
(d) Results using GPT-4
\vspace{-0.15cm}

\caption{Results of $\text{ByCS}_\text{fast}$ on speech and text tasks. Results of best baseline and ByCS are also shown for comparison.}
\label{t14}
\end{table*}

\end{document}